\pdfoutput=1

\documentclass[11pt]{article}

\usepackage[]{acl}

\usepackage{amsmath}
\usepackage{times}
\usepackage{latexsym}
\usepackage{mathabx}
\usepackage{algorithm}
\usepackage{algpseudocode}

\usepackage[T1]{fontenc}

\usepackage[utf8]{inputenc}

\usepackage{microtype}

\usepackage{makecell}
\usepackage{booktabs}
\usepackage{graphicx}
\usepackage{multirow}
\usepackage[labelfont=bf, format=plain, labelformat=simple, font=small, compatibility=false]{caption}

\title{Rule By Example: Harnessing Logical Rules for Explainable Hate Speech Detection}

\author{
  Christopher Clarke$^{*\dagger}$\hspace{10pt} Matthew Hall$^\ddagger$\hspace{10pt} Gaurav Mittal$^\ddagger$\hspace{10pt} Ye Yu$^\ddagger$\hspace{10pt} \\  \textbf{Sandra Sajeev$^\ddagger$\hspace{10pt}  Jason Mars$^\dagger$\hspace{10pt} Mei Chen$^\ddagger$\hspace{10pt}}    \vspace{0.3cm}\\
    \text{$^\dagger$University of Michigan, Ann Arbor, MI}\\
    \text{$^\ddagger$Microsoft, Redmond, WA}\\
    \text{\{csclarke, profmars\}@umich.edu} \\
    \text{\{mathall, gaurav.mittal, yu.ye,  ssajeev, mei.chen\}@microsoft.com}
}

\begin{document}
\maketitle
\begin{abstract}

Classic approaches to content moderation typically apply a rule-based heuristic approach to flag content.
While rules are easily customizable and intuitive for humans to interpret, they are inherently fragile and lack the flexibility or robustness needed to moderate the vast amount of undesirable content found online today.
Recent advances in deep learning have demonstrated the promise of using highly effective deep neural models to overcome these challenges. However, despite the improved performance, these data-driven models lack transparency and explainability, often leading to mistrust from everyday users and a lack of adoption by many platforms.
In this paper, we present \textbf{Rule By Example} (RBE): a novel exemplar-based contrastive learning approach for learning from logical rules for the task of textual content moderation. RBE is capable of providing rule-grounded predictions, allowing for more explainable and customizable predictions compared to typical deep learning-based approaches. We demonstrate that our approach is capable of learning rich rule embedding representations using only a few data examples.
Experimental results on 3 popular hate speech classification datasets show that RBE is able to outperform state-of-the-art deep learning classifiers as well as the use of rules in both supervised and unsupervised settings while providing explainable model predictions via rule-grounding. 

\end{abstract}

\begingroup\def\thefootnote{*}\footnotetext{This work was done as Christopher’s internship project at Microsoft.}\endgroup

\section{Introduction}

Content moderation is a major challenge confronting the safety of online social platforms such as Facebook, Twitter, YouTube, Twitch, etc. \cite{9623288}. Major technology corporations are increasingly allocating valuable resources towards the development of automated systems for the detection and moderation of harmful content in addition to hiring and training expert human moderators to combat the growing menace of negativity and toxicity online \cite{wagner_bloomberg_2021, wharton}.

\begin{figure}
  \centering
    \includegraphics[width=\columnwidth]{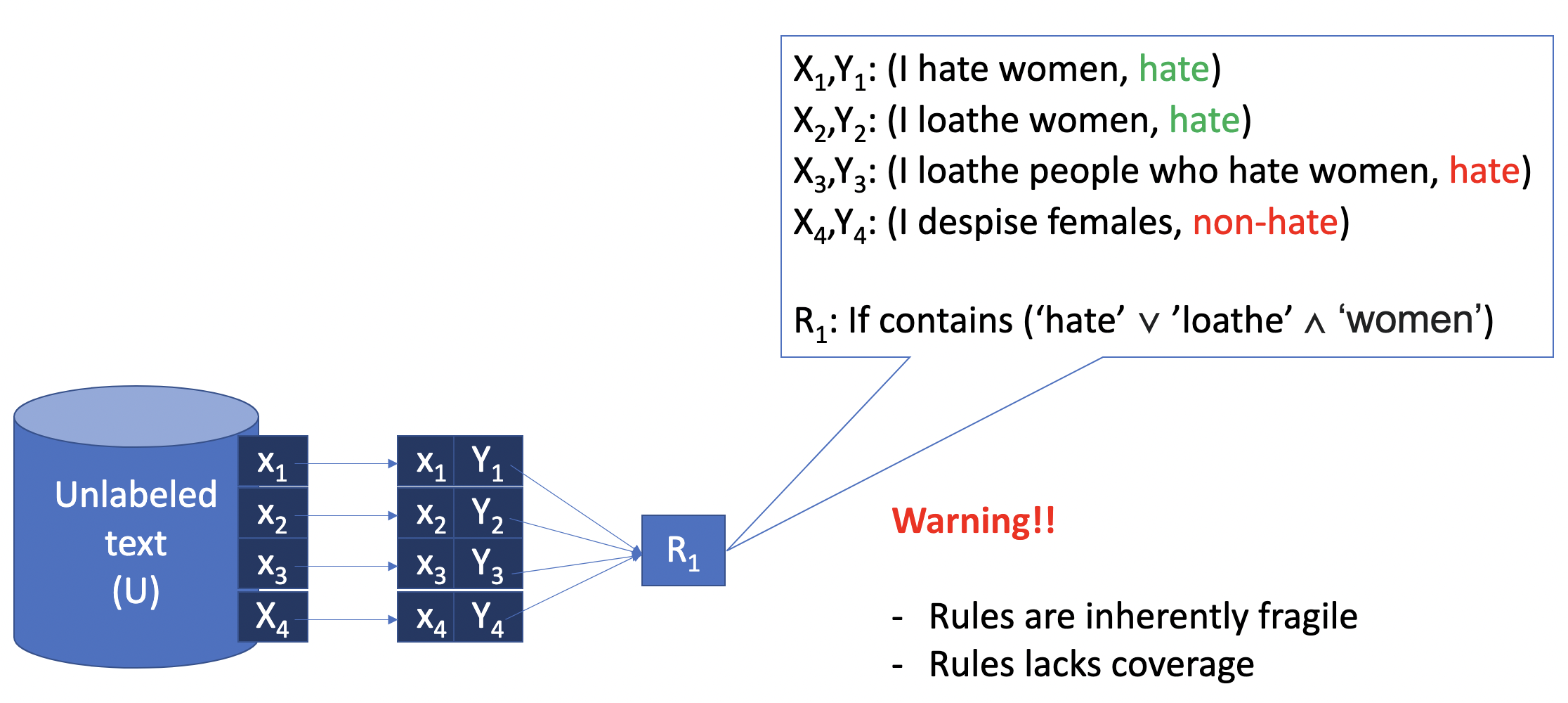}
    \caption{Generalization problem of rules. Logical rules, while easy to explain, are inherently fragile to the nuances of natural language.}
  \label{fig:problem}
  \vspace{-1.8pc}
\end{figure}

Despite the popularity of deep learning approaches, many practical solutions used in products today are comprised of rule-based techniques based on expertly curated signals such as block lists, key phrases, and regular expressions \cite{gillespie2018custodians, zhang_2019, DADA2019e01802}. Such methods are widely used due to their transparency, ease of customization, and interpretability. However, they have the disadvantage of being difficult to maintain and scale, in addition to being inherently fragile and noisy \cite{zhang_2019, davidson_2017, lee_2022, lai_2022}. Figure \ref{fig:problem} shows an example where logical rules, while explainable in nature, face the problem of being inflexible to their context of use in natural language. While a given rule may be too specific and fail to capture different variations of usage commonly found in content online, rules can also be too broad and incorrectly block lexically similar content.
 
In contrast to the challenges faced by rule-based methods, data-driven deep learning approaches have shown great promise across a wide range of content moderation tasks and modalities \cite{deep_learning_hate_2022, shido-etal-2022-textual, lai_2022}. Fueled by large amounts of data and deep neural networks, these complex models are capable of learning richer representations that better generalize to unseen data. The impressive performances of these models have resulted in significant industry investment in content moderation as-a-service. Several technology companies such as Google \footnote{\url{https://perspectiveapi.com/}}, OpenAI \footnote{\url{https://openai.com/blog/new-and-improved-content-moderation\%2Dtooling/}}, and Microsoft \footnote{\url{https://azure.microsoft.com/en-us/products/cognitive-services/content-moderator/}} use these models to offer services to aid in content moderation. However, despite their significant investment, they face adoption challenges due to the inability of customers to understand how these complex models reason about their decisions \cite{tarasov_2021, haimson_2021, juneja_2020}. Additionally, with the increasing attention around online content moderation and distrust amongst consumers, explainability and transparency are at the forefront of demands \cite{kemp_ekins_2021, hate_alert_das_mathew_saha_2022}. This presents the challenging open question of how we can leverage the robustness and predictive performance of complex deep-learning models whilst allowing the transparency, customizability, and interpretability that rule-based approaches provide.

Prior works such as \citet{awasthi_2020, sunyong_2021, reid_2022} have explored learning from rules for tasks such as controlling neural network learning, assisting in human annotation, and improving self-supervised learning in low data scenarios. \citet{awasthi_2020} propose a rule-exemplar training method for noisy supervision using rules. While performant in denoising over-generalized rules in the network via a soft implication loss, similar to other ML approaches, this method lacks the ability to interpret model predictions at inference time. \citet{reid_2022} propose a general-purpose framework for the automatic discovery and integration of symbolic rules into pre-trained models. However, these symbolic rules are derived from low-capacity ML models on a reduced feature space. While less complex than large deep neural networks, these low-capacity models are still not easily interpretable by humans. Therefore, the task of combining the explainability of rules and the predictive power of deep learning models remains an open problem.

In order to tackle this problem, we introduce \textbf{Rule By Example} (RBE): a novel exemplar-based contrastive learning approach for learning from logical rules for the task of textual content moderation. RBE is comprised of two neural networks, a rule encoder, and a text encoder, which jointly learn rich embedding representations for hateful content and the logical rules that govern them. Through the use of contrastive learning, our framework uses a semantic similarity objective that pairs hateful examples with clusters of rule exemplars that govern it. Through this approach, RBE is able to provide more explainable predictions by allowing for what we define as \textit{Rule-grounding}. This means that our model is able to ground its predictions by showing the corresponding explainable logical rule and the exemplars that constitute that rule.

We evaluate RBE in both supervised and unsupervised settings using a suite of rulesets.
Our results show that with as little as one exemplar per rule, RBE is capable of outperforming state-of-the-art hateful text classifiers across three benchmark content moderation datasets in both settings. In summary, the contributions of this paper are:

\begin{itemize}
    \item Rule By Example (RBE): a novel exemplar-based contrastive learning approach to learn from logical rules for the task of textual content moderation.\footnote{\url{https://github.com/ChrisIsKing/Rule-By-Example}}

    \item We demonstrate how RBE can be easily integrated to boost model F1-score by up to 4\% on three popular hate speech classification datasets.

    \item A detailed analysis and insights into the customizability and interpretability features of RBE to address the problem of emerging hateful content and model transparency.
\end{itemize}

\begin{figure*}[t]
    \centering
    \includegraphics[width=1\textwidth]{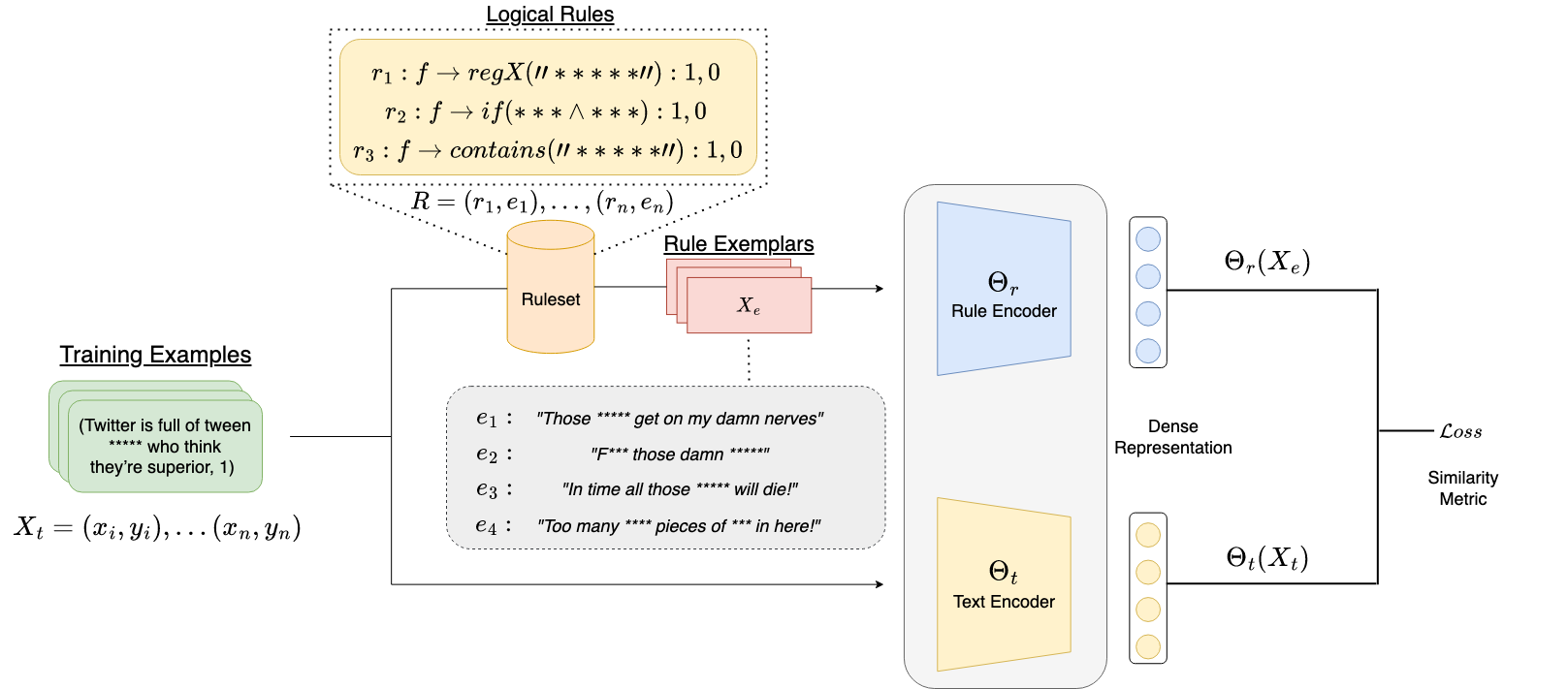}
    \caption{\textbf{Rule By Example Framework}:  RBE is comprised of two neural networks, a rule encoder and a text encoder, which jointly learn rich embedding representations for hateful content and the logical rules that govern them. Through Contrastive learning, RBE utilizes a semantic similarity objective that pairs hateful examples with clusters of rule exemplars that govern it.}
    \label{fig:model}
     \vspace{-1pc}
\end{figure*}

\section{Rule By Example Framework}
In this section, we outline the Rule By Example framework, define its operational terms, and describe its end-to-end architecture.
We first formally describe the two main operational terms used in our framework: 1) \textbf{Ruleset} - a ruleset is comprised of a series of executable functions that when given text as input ``fire'' if and only if all conditions defined in the rule are met by the input. Figure \ref{fig:problem} shows an example of a simple rule that is triggered if a given text contains the keywords \textit{``hate''} or \textit{``loathe''} and contains \textit{``women''}. Rules can be any programmable function that acts on text such as regular expressions, blocklists, keywords, etc. In the scope of this work, we only consider simple rules that humans can easily interpret. As such an ML model cannot be considered a rule, given their black-box nature. 2) \textbf{Exemplar} - an exemplar is a given textual example that well-defines the type of content governed by a rule. For example, $X_1$ and $X_2$ in Figure \ref{fig:problem} can be considered exemplars of rule $R_1$ since they correctly match the conditions of $R_1$.                            

Consider a ruleset of rule-exemplar pairs $R {=} \{(r_1, e_1), (r_2, e_2),...,(r_n, e_n)\}$ where $r_i$ denotes a defined rule and $e_i$ denotes an exemplar for which $r_i$ correctly fires. For a given corpus $X$ comprising labeled examples $X {=} \{(x_1,y_1), (x_2, y_2),...,(x_m,y_m)\}$, each rule $r_i$ can be used as a black-box function $R_i: x \rightarrow \{y_i, \emptyset \}$ to noisily label each instance $x$ such that it assigns a label $y$ or no label at all. An instance may be covered by more than one rule or no rule at all. Additionally, the cover set $C$ denotes the set of instances in $X$ where a rule $r_i$ fires. The generalization problem that arises when rules are applied noisily is two-fold. When rules are too broad the cover set $C$ is large and incorrectly labels a large amount of non-hateful content. Likewise, when rules are too strict and fragile, the cover set $C$ is too small, and lexically and semantically similar content that is hateful ends up being ignored. Our goal is to leverage these rules and their exemplars to facilitate explainable model learning.

\begin{algorithm}
\small
\caption{Supervised Dual Encoder Training}\label{alg:rank}
\textbf{Require:} Rule Encoder $\Theta_r$ Text Encoder $\Theta_t$
\begin{algorithmic}[1]
\Require{Training Data $X{=} (x_1,y_1)...(x_n,y_n)$, Ruleset $R {=} (r_1, e_1),...,(r_n, e_n)$}
\Ensure{Updated parameters $\Theta_r$, $\Theta_t$}
\State Initialize $\Theta_r$ and $\Theta_t$
\While {not converged}
    \State Get mini-batch $X_b$
    \For {each instance $x_i$ in $X_b$}
        \State Get exemplars $e_i = doRuleset(R, x_i)$
        \State Concatenate exemplars $e_i$
    \EndFor
    \State Get $\Theta_r(E_b)$ and $\Theta_t(X_b)$
    \State Compute $\mathcal{L} {=} \frac{1}{2}(Y_bD^2+(1-Y_b)max(margin - D, 0)^2)$
    \State Update parameters of $\Theta_r$ and $\Theta_t$
\EndWhile
\end{algorithmic}
\end{algorithm}

\subsection{Dual Encoder Architecture}
The Dual-Encoder architecture, as
illustrated in Figure \ref{fig:model}, is commonly used in dense retrieval systems and multi-modal applications \cite{clarke-etal-2022-one, sbert, laprador}. Our architecture consists of a Rule Encoder $\Theta_r$ and a Text Encoder $\Theta_t$. These are two Bert-like bidirectional transformer models \cite{bert} each responsible for learning embedding representations of their respective inputs.
This Dual Encoder architecture enables pre-indexing of exemplars allowing for faster inference at runtime after training.

\paragraph{Encoding Pipeline}~Given an input text $x_t$, we first extract the set of applicable rules and their respective exemplars from the ruleset $R$. We then concatenate each extracted exemplar to form $x_e$. In the event that no rules are applicable to $x_t$, we randomly sample exemplars from the entire ruleset to form $x_e$. Using the form $x_e = \left \{\left [CLS\right], e^1_{1},...,e^1_{m},\left[SEP\right], e^n_{1},....,e^n_{k}    \right \}$, we then use rule encoder $\Theta_r$ to encode $x_e$ into hidden states $h_e= \left \{v_{[CLS]},v_1,...,v_{[SEP]} \right \}$ where $e^n_k$ is the $k$-th token of the $n$-th exemplar and $[SEP]$ and $[CLS]$ are special tokens. Similarly, using the text encoder $\Theta_t$, we encode $x_t$. In order to obtain a dense representation, we apply a mean pooling operation to the hidden states and derive a fixed-sized sentence embedding. After obtaining the representation for both the exemplars $x_e$ and the text $x_t$, we use the cosine function to measure the similarity between them:
\begin{equation}
    sim(x_e,x_t) = \frac{\Theta_r(x_e)\cdot\Theta_t(x_t)}{\left \| \Theta_r(x_e) \right \|\left \| \Theta_t(x_t) \right \|}
\end{equation}
We employ a contrastive loss \cite{contrastive} to learn the embedding representations for our rule and text encoder. Contrastive learning encourages the model to maximize the representation similarity between \textit{same-label} examples and to minimize it for \textit{different-label} examples. This enables the embedding representations of our encoded ruleset to match the representation of the text correctly covered by cover set $C$. Likewise, for benign examples that rules incorrectly cover, our contrastive learning objective increases the distance between those representations, thus restricting the over-generalization of certain rules in the ruleset. Let $Y_t$ be the correct label of the texts $X_t$, $D$ be the cosine distance of $(x_e, x_t)$ and $m$ be the margin, our contrastive learning loss function is defined as follows:
\begin{equation}
\small
    \mathcal{L} = \frac{1}{2}(Y_tD^2+(1-Y_t)max(m - D, 0)^2)
\end{equation}
The training loop, with the encoding pipeline and constrastive loss step, are detailed in Algorithm \ref{alg:rank}.

\subsection{Rule-Grounding} \label{sec:rule_grounding}
By taking an embeddings-based approach to learning representations, RBE enables what we define as \textit{rule-grounding}. Rule-grounding enables us to trace our model predictions back to the explainable ruleset accompanied by the exemplars that define each rule. For any input $x_t$ that has been marked as positive by our dual encoder, we perform a rules search to find which rules fire on that input as well as an embedding similarity search to find the nearest exemplars and the rules those exemplars belong to. Table \ref{tab:rule_grounding} shows an example of this.

\section{Experimental Setup}

\paragraph{Training} We train all models with AdamW optimizer and weight decay of 0.01 on all data. We employ early stopping with a ceiling of 10 epochs, a learning rate of $2\text{e-}5$, batch size of 8, and linear learning rate warmup over the first 10\% steps with a cosine schedule. Our models are trained with NVIDIA Tesla V100 32GB GPUs using Azure Machine Learning Studio. We pre-process data and train all models with different random seeds over multiple runs. Our implementation of RBE is based on Huggingface Transformers \cite{wolf-etal-2020-transformers} and Sentence Transformers \cite{sbert}. RBE utilizes two Bert-based networks consisting of 110 million parameters each. Approximately 2,000 GPU hours were required to train all hyperparameter variations of RBE plus the Bert baseline across all 3 test sets.

\paragraph{Baselines} We evaluate our training algorithms in both supervised and unsupervised settings. We compare against the baselines of applying logical rules as is and the current SOTA approach of training transformer-based sequence classifiers \cite{hatexplain}.

\subsection{Datasets}
We evaluate RBE across three datasets on the task of hate-speech classification. Across each dataset, we frame the problem as a binary classification task of detecting whether a given text is hateful  or non-hateful. We augment each dataset with rulesets that we manually curate. More information on each dataset and ruleset is provided below.

\paragraph{HateXplain} \cite{hatexplain} is a large-scale benchmark dataset for explainable hate speech detection that covers multiple aspects of hate speech detection. It consists of $\sim$20k samples across 3 labels ``hateful'', ``offensive'', and ``normal''. Additionally, each sample is accompanied by a corresponding target group and explainable rationales. In our experiments, we combine the output classes of hateful and offensive into one resulting in $\sim$8k/1k/1k hateful samples and $\sim$6k/781/782 non-hateful samples for train/validation/test respectively. Additionally, we utilize the accompanying rationales for ruleset construction.

\paragraph{Jigsaw\footnote{\url{https://www.kaggle.com/competitions/jigsaw-toxic-comment-classification\%2Dchallenge}}} is a large-scale dataset of Wikipedia comments labeled by human raters for toxic behavior. The defined types of toxicity are ``toxic'', ``severe toxic'', ``obscene'', ``threat'', ``insult'', and ``identity hate''. Each comment can have any one or more of these labels. In total, it contains $\sim$230k samples. In our experiments, we define examples of the ``identity hate'' class as hateful and the rest as non-hateful resulting in a dataset of 1405/100/712 hateful samples and $\sim$158k/1k/63k non-hateful examples for train/validation/test respectively.

\paragraph{Contextual Abuse Dataset (CAD)} \cite{cad_2021} is annotated dataset of $\sim$25k Reddit entries labeled across six conceptually distinct primary categories of ``Identity-directed'', ``Person-directed'', ``Affiliation directed'', ``Counter Speech'', ``Non-hateful Slurs'', and ``Neutral''. In our experiment, we define examples of the ``identity-directed'' class as hateful and treat the remaining examples as non-hateful resulting in a dataset of 1353/513/428 hateful samples and $\sim$12k/4k/4k non-hateful samples for train/validation/test.

\subsection{Ruleset Construction} \label{sec:ruleset_construction}
\paragraph{Hate+Abuse List} We utilize a ruleset targeting identity hate which we'll refer to as \textbf{Hate+Abuse List}. It consists of a list of n-grams representing harmful language such as slurs or hate verbs. Hate+Abuse List is similar to the publically available bad word lists commonly found online. We treat each n-gram entry in Hate+Abuse List as its own rule that proposes a positive label if the n-gram is in the input text. In total, Hate+Abuse List consists of 2957 distinct identity hate rules.

\paragraph{HateXplain Rationale Ruleset} Using the labeled annotator rationales included in the HateXplain dataset, we programmatically generate a Ruleset for HateXplain. To do so, we extract 1, 2, and 3-gram substrings from the annotator rationales and cluster them by annotator-identified target demographic groups. We then take the top N n-grams per each demographic group and automatically create rules for each of them. This results in rules similar in nature to our Hate+Abuse List. Using a default cluster size of 100 across the 25 target categories defined in HateXplain, we generated a total of 670 distinct rules for HateXplain.

\paragraph{Contextual Abuse Rationale Ruleset} Similar to our derived HateXplain ruleset we programmatically generate a Ruleset for the Contextual Abuse Dataset using annotator-labeled rationales. Following the identical process outlined before, this results in a total of 2712 distinct rules for CAD.

\paragraph{Exemplar Selection} For each dataset we complete our Ruleset construction by pairing each rule with accompanying exemplars. To achieve this, we first run our Ruleset on the dataset trainset and extract instances for which a rule correctly fires. For each rule that correctly fires, we then randomly select N instances to act as the exemplars. Additionally, to restrict potentially overgeneralized rules we enforce the condition that no two rules can be mapped to the same exemplar. Unless stated otherwise, we report results using just one exemplar per rule in our experiments. 

\begin{table*}[]
    \resizebox{\textwidth}{!}{%
    \begin{tabular}{|lllllclllclll|}
    \hline
    \multicolumn{13}{|c|}{Content Moderation Using Rules (Fully Supervised)}                                                                                                                                                                                                                                                                                                                                                                                                                                                      \\ \hline
    \multicolumn{1}{|l|}{}                                    & \multicolumn{4}{c|}{HateXplain}                                                                                                                       & \multicolumn{4}{c|}{Jigsaw}                                                                                                                           & \multicolumn{4}{c|}{CAD}                                                                                                                          \\ \hline
    \multicolumn{1}{|l|}{\textbf{Model}}                      & \multicolumn{1}{c|}{Precision}      & \multicolumn{1}{c|}{Recall}         & \multicolumn{1}{c|}{F1}             & \multicolumn{1}{c|}{Acc}            & \multicolumn{1}{c|}{Precision}      & \multicolumn{1}{c|}{Recall}         & \multicolumn{1}{c|}{F1}             & \multicolumn{1}{c|}{Acc}            & \multicolumn{1}{c|}{Precision}      & \multicolumn{1}{c|}{Recall}         & \multicolumn{1}{c|}{F1}             & \multicolumn{1}{c|}{Acc}        \\ \hline
    \multicolumn{1}{|l|}{HateXplain Rules}          & \multicolumn{1}{c|}{0.609}          & \multicolumn{1}{c|}{0.983} & \multicolumn{1}{c|}{0.752}          & \multicolumn{1}{c|}{0.615}          & \multicolumn{1}{c|}{-}              & \multicolumn{1}{c|}{-}              & \multicolumn{1}{c|}{-}              & \multicolumn{1}{c|}{-}              & \multicolumn{1}{c|}{-}              & \multicolumn{1}{c|}{-}              & \multicolumn{1}{c|}{-}              & \multicolumn{1}{c|}{-}          \\ \hline
    \multicolumn{1}{|l|}{Hate+Abuse Rules}                    & \multicolumn{1}{c|}{0.755}          & \multicolumn{1}{c|}{0.687}          & \multicolumn{1}{c|}{0.719}          & \multicolumn{1}{c|}{0.682}          & \multicolumn{1}{c|}{0.164}          & \multicolumn{1}{c|}{0.361}          & \multicolumn{1}{c|}{0.226}          & \multicolumn{1}{c|}{0.972}          & \multicolumn{1}{c|}{0.586}          & \multicolumn{1}{c|}{0.193}          & \multicolumn{1}{c|}{0.290}          & 0.909                  \\ \hline
    \multicolumn{1}{|l|}{CAD Rules}          & \multicolumn{1}{c|}{-}          & \multicolumn{1}{c|}{-} & \multicolumn{1}{c|}{-}          & \multicolumn{1}{c|}{-}          & \multicolumn{1}{c|}{-}              & \multicolumn{1}{c|}{-}              & \multicolumn{1}{c|}{-}              & \multicolumn{1}{c|}{-}              & \multicolumn{1}{c|}{0.110}              & \multicolumn{1}{c|}{0.842}              & \multicolumn{1}{c|}{0.194}              & \multicolumn{1}{c|}{0.325}          \\ \hline
    \multicolumn{1}{|l|}{BERT$^+$}                        & \multicolumn{1}{c|}{0.808}          & \multicolumn{1}{c|}{0.841}          & \multicolumn{1}{c|}{0.824}          & \multicolumn{1}{c|}{0.787} & \multicolumn{1}{c|}{0.459}          & \multicolumn{1}{c|}{0.729}          & \multicolumn{1}{c|}{0.563}          & \multicolumn{1}{c|}{0.987}          & \multicolumn{1}{c|}{0.445}          & \multicolumn{1}{c|}{0.421}          & \multicolumn{1}{c|}{0.433}          & 0.893                           \\ \hline
    \multicolumn{1}{|l|}{MPNet$^\wedge$}                        & \multicolumn{1}{c|}{0.795}          & \multicolumn{1}{c|}{0.854}          & \multicolumn{1}{c|}{0.823}          & \multicolumn{1}{c|}{0.783} & \multicolumn{1}{c|}{0.510}          & \multicolumn{1}{c|}{0.674}          & \multicolumn{1}{c|}{0.581}          & \multicolumn{1}{c|}{0.989}          & \multicolumn{1}{c|}{0.519}          & \multicolumn{1}{c|}{0.417}          & \multicolumn{1}{c|}{0.463}          & 0.906                           \\ \hline
    \multicolumn{1}{|l|}{Rule By Example$^{+\triangle}$}  & \multicolumn{1}{c|}{0.758} & \multicolumn{1}{c|}{0.903}          & \multicolumn{1}{c|}{0.824}          & \multicolumn{1}{c|}{0.771}          & \multicolumn{1}{c|}{0.581} & \multicolumn{1}{c|}{0.625}          & \multicolumn{1}{l|}{0.602} & \multicolumn{1}{c|}{0.991} & \multicolumn{1}{c|}{0.416}          & \multicolumn{1}{c|}{0.478} & \multicolumn{1}{c|}{0.445} & 0.885                  \\ \hline
    \multicolumn{1}{|l|}{Rule By Example$^{\wedge\triangle}$} & \multicolumn{1}{c|}{0.790}          & \multicolumn{1}{c|}{0.891}          & \multicolumn{1}{c|}{\textbf{0.837}}          & \multicolumn{1}{c|}{0.795}          & \multicolumn{1}{c|}{0.508}          & \multicolumn{1}{c|}{0.746} & \multicolumn{1}{c|}{\textbf{0.604}}          & \multicolumn{1}{c|}{0.989}          & \multicolumn{1}{c|}{0.484} & \multicolumn{1}{c|}{0.468}          & \multicolumn{1}{l|}{0.476} & 0.900                           \\ \hline
    \multicolumn{1}{|l|}{Rule By Example$^{+\ast}$}  & \multicolumn{1}{c|}{0.738}          & \multicolumn{1}{c|}{0.912}          & \multicolumn{1}{c|}{0.816}          & \multicolumn{1}{c|}{0.756}          & \multicolumn{1}{c|}{-}              & \multicolumn{1}{c|}{-}              & \multicolumn{1}{c|}{-}              & \multicolumn{1}{c|}{-}              & \multicolumn{1}{c|}{-}              & \multicolumn{1}{c|}{-}              & \multicolumn{1}{c|}{-}              & \multicolumn{1}{c|}{-} \\ \hline
    \multicolumn{1}{|l|}{Rule By Example$^{\wedge\ast}$} & \multicolumn{1}{c|}{0.779}          & \multicolumn{1}{c|}{0.893}          & \multicolumn{1}{l|}{0.832} & \multicolumn{1}{c|}{0.786}          & \multicolumn{1}{c|}{-}              & \multicolumn{1}{c|}{-}              & \multicolumn{1}{c|}{-}              & \multicolumn{1}{c|}{-}              & \multicolumn{1}{c|}{-}              & \multicolumn{1}{c|}{-}              & \multicolumn{1}{c|}{-}              & \multicolumn{1}{c|}{-}          \\ \hline
    \multicolumn{1}{|l|}{Rule By Example$^{+\ddagger}$}  & \multicolumn{1}{c|}{-}          & \multicolumn{1}{c|}{-}          & \multicolumn{1}{c|}{-}          & \multicolumn{1}{c|}{-}          & \multicolumn{1}{c|}{-}              & \multicolumn{1}{c|}{-}              & \multicolumn{1}{c|}{-}              & \multicolumn{1}{c|}{-}              & \multicolumn{1}{c|}{0.512}              & \multicolumn{1}{c|}{0.378}              & \multicolumn{1}{c|}{0.435}              & \multicolumn{1}{c|}{0.905} \\ \hline
    \multicolumn{1}{|l|}{Rule By Example$^{\wedge\ddagger}$} & \multicolumn{1}{c|}{-}          & \multicolumn{1}{c|}{-}          & \multicolumn{1}{l|}{-} & \multicolumn{1}{c|}{-}          & \multicolumn{1}{c|}{-}              & \multicolumn{1}{c|}{-}              & \multicolumn{1}{c|}{-}              & \multicolumn{1}{c|}{-}              & \multicolumn{1}{c|}{0.508}              & \multicolumn{1}{c|}{0.448}              & \multicolumn{1}{c|}{\textbf{0.476}}              & \multicolumn{1}{c|}{0.905}          \\ \hline
    \end{tabular}%
    }
    \caption{Experiment Results in Fully Supervised Setting on hate speech classification datasets. $^+$Uses BERT \cite{bert} as the base model. $^\wedge$Uses MPNet \cite{mpnet} as the base model. $^\ast$Uses HateXplain ruleset. $^\triangle$Uses Hate+Abuse ruleset. $^\ddagger$Uses CAD Ruleset. \textbf{Note:} The HateXplain Ruleset and Contextual Abuse Dataset (CAD) Ruleset are only applicable to their respective datasets.}
    \label{tab:supervised}
    \vspace{-1pc}
    \end{table*}

\subsection{Unsupervised Setting}
In addition to evaluating RBE in supervised settings, we investigate the applicability of RBE in unsupervised settings where no labeled data is present. In this setting, we are presented with a large unlabeled corpus $T$ and a given ruleset $R$. This setting is particularly challenging due to the inherent generalization problem of rules. Loosely applying rules as is in this setting results in the model overfitting to the distribution of the ruleset as seen in Table \ref{tab:unsupervised}. To combat this issue, we design three different semantic clustering-based strategies for determining rule quality in an unsupervised setting: \textit{Mean}, \textit{Concat}, and \textit{Distance} clustering. Given an unlabeled corpus $T=\left\{ t_1, t_2,...,t_n\right \}$, ruleset $R = \{(r_1, e_1),...,(r_n, e_n)\}$, and a threshold $k$, we first encode the entire corpus $T$ using a pre-trained sentence embedding model $E_\Theta$. In our case, we use a fine-tuned version of MPNet \cite{mpnet} from the Sentence Transformers library. After receiving our encoded corpus $E_\Theta(T)$, for the \textit{Mean} and \textit{Concat}, we construct a rule embedding $r^i_\Theta$ for each rule $r_i$ in the ruleset. In the \textit{Mean} strategy, this is obtained by taking the mean of all rule exemplars $\mu(r^i_\Theta) = ( \frac{1}{m} \sum_{i}^{m} e^i_{m} )$. For \textit{Concat}, this is calculated by concatenating all rule exemplars $\mu(r_i) = E_\Theta(e^i_1 \mathbin\Vert ... \mathbin\Vert e^i_m)$ and encoding the concatenated representation. Once $r^i_\Theta$ is constructed, we then label each text in the corpus whose cosine similarity is within the threshold $k$:

\begin{equation}
    f(t_i)= 
\begin{cases}
    1,& \text{if } sim(r^i_\Theta, E_\Theta(t_i))\geq k\\
    0,              & \text{otherwise}
\end{cases}
\end{equation}

In contrast to the \textit{Mean} and \textit{Concat} strategies, the \textit{Distance} strategy takes a rule elimination approach. Given an unlabeled corpus $T=\left\{ t_1, t_2,...,t_n\right \}$, ruleset $R = \{(r_1, e_1),...,(r_n, e_n)\}$, and a threshold $k$, we first noisily label the entire corpus using the ruleset $R_i: x_t \rightarrow \{1, \emptyset \}$ such that each rule is paired with a cover set $R = \{(r_1, e_1, c_1),...,(r_n, e_n, c_n)\}$ where $c_i$ is the set of texts in covered by $r_i$. Next, for each rule, we encode text in its cover set $E_\Theta(c_i)$ and calculate the average cosine distance between each embedding and its neighboring examples in $c_i$.

\begin{equation}
    avgDist(E_\Theta(c_i)) = \frac{1}{n} \sum_{i}^{n} dist(c^i_{j}, c^i_{j-1} )
\end{equation}

Lastly, once the average distance for each rule is calculated, using the defined threshold $k$, we flip any weakly labeled examples in the cover set if the average distance for that rule is above the threshold $k$:

\begin{equation}
    f(t_i)= 
\begin{cases}
    1,& \text{if } avgDist(r_i)\geq k\\
    0,              & \text{otherwise}
\end{cases}
\end{equation}

\begin{table*}[]
\resizebox{\textwidth}{!}{%
\begin{tabular}{|llp{1cm}p{12cm}|}
\hline
\multicolumn{4}{|c|}{\textbf{Rule By Example Rule Grounding}} \\
\hline
\multicolumn{1}{|l|}{\textbf{Dataset}} & \multicolumn{1}{c|}{\textbf{Text}} & \multicolumn{1}{c|}{\textbf{Fired Rules}} & \multicolumn{1}{c|}{\textbf{Exemplar}} \\
\hline
\multicolumn{1}{|l|}{HateXplain} & \multicolumn{1}{p{6cm}|}{fully agree every personal interaction with these queers reinforces what i already knew severe mental illness and obnoxious to boot} & \multicolumn{1}{p{4cm}|}{if contains("queers") $\rightarrow 1,0$} & yes but queers are too self righteous to let them be included originally the gay and pedophile communities were working together nambla was started by gay men who liked boys now all subsets of pedophiles are members assuming nambla still exists \\
\hline
\multicolumn{1}{|l|}{Jigsaw} & \multicolumn{1}{p{6cm}|}{Why do they put so many gay people on the damn show since when it was okay to be gay.} & \multicolumn{1}{l|}{if contains("so" $\land$ "gay") $\rightarrow 1,0$} & stop reverting my edit your so f**** gay get a f**** life your f**** or go get laid or something \\
\hline
\multicolumn{1}{|l|}{CAD} & \multicolumn{1}{p{6cm}|}{What a little b****} & \multicolumn{1}{l|}{if contains("b****") $\rightarrow 1,0$} & Nope, today is tuna b**** \\
\hline
\end{tabular}%
}
\caption{Example of Rule-grounding explanations. This table shows examples of traced model predictions produced by RBE. By displaying the rules and exemplars responsible, rule authors and users are better able to understand model predictions and can automatically adjust their ruleset to further improve model performance. }
\label{tab:rule_grounding}
\vspace{-1pc}
\end{table*}

\section{Results and Discussion}
We analyze the results of our experiments, detail our insights, and discuss the implications of applying RBE for explainable hate speech detection.

\paragraph{Evaluation Metrics:} The precision, recall, and F1 score for each dataset in a supervised setting are reported in Table \ref{tab:supervised}. Due to the highly skewed class distribution, we favor macro F1 scores as our main evaluation metric. We also report accuracy scores (the fraction of entries for which the full set of labels matches) as another metric.

\subsection{Supervised Performance}
Table \ref{tab:supervised} reports our results on three hate speech classification datasets in the supervised setting. We observe that RBE is able to outperform SOTA transformer-based models BERT and MPNet by 1.3/1.4\%, 4.1/2.3\%, and 4.3/1.3\% in F1-score on HateXplain, Jigsaw, and CAD respectively. This improvement highlights the impact of leveraging rules in the training process of our framework. Additionally, it is important to note that this increase was achieved using only 1 exemplar per rule in the ruleset. These exemplars were also used to train the comparative baseline models, ensuring that all approaches were trained on the same number of samples. This further showcases how lightweight and flexible RBE is to integrate into a content moderation workflow.
For HateXplain, our experiments show that the combination of MPNet as the initialized encoder with both the HateXplain Rationale and Hate+Abuse Ruleset delivers the best performance. Upon deeper analysis, we find that this is due to two main factors: 

1) \textbf{Ruleset Size and Alignment} - As explained in Section \ref{sec:ruleset_construction} the HateXplain Rationale Ruleset was automatically crafted using rationale labels from expert annotators. This results in a powerful ruleset capable of identifying a large amount of hateful content in the HateXplain dataset as shown by the high recall score of the HateXplain Rationale Ruleset in Table \ref{tab:supervised}. Additionally, when applied to the HateXplain dataset, the HateXplain Rationale Ruleset produces a total of 577 rules compared to the 377 rules derived from the Hate+Abuse Ruleset, allowing for more rule representations for the model to contrast against.

2) \textbf{Embedding Initialization} - Out of the box, pre-trained BERT does not produce meaningfully distinct sentence representations. In practice, the BERT [CLS] token as well as averaged BERT outputs can contain useful information after downstream fine-tuning. This is shown by the BERT performance in Table \ref{tab:supervised}. However, when the pre-trained model output is pooled across all dimensions and used for calculating semantic similarity, this results in similar representations even for completely different input text. As a result, if applied to the HateXplain dataset without any fine-tuning, BERT embeddings obtain a precision, recall, and F1-score of 59\%, 100\%, and 75\% respectively, where every example is labeled as hateful. This lack of varied sentence representation coupled with a verbose ruleset such as the HateXplain Rationale Ruleset results in an initial biasing towards hateful examples as shown by the high recall scores. As such, utilizing a pre-trained sentence embedder, such as MPNet, with a pre-train task more optimized for semantic embeddings results in better performance. We observe a similar trend when utilizing our derived ruleset for CAD. \textbf{Note:} When trained longer, the bias of the BERT model decreases as more varied sentence representations are learned.

On Jigsaw and Contextual Abuse datasets using the Hate+Abuse List and derived CAD Ruleset, RBE outperforms SOTA by an increased margin of 4.1/2.3\%, and 4.3/1.3\% respectively. Contrary to HateXplain, these two datasets are more heavily imbalanced toward non-hateful examples and thus more representative of the real-world case of content moderation where most content is considered benign. This increased performance highlights the power of incorporating logical rules to assist model learning and also the ability of RBE to better generalize rules. As seen in Table \ref{tab:supervised}, on its own the Hate+Abuse ruleset performs poorly on each dataset in both precision and recall. Despite RBE's reliance on this ruleset to guide model learning, when combined with labeled training data, RBE is capable of both restricting over-generalized rules and leveraging its understanding of semantic similarity to extend fragile rules regardless of the base model. Additionally, when using the CAD ruleset which is heavily overfitted to the CAD dataset, as shown by the skewed recall score, RBE is still capable of outperforming the baselines.

\paragraph{Out-of-domain Rulesets} Our Hate+Abuse ruleset is a generic ruleset unrelated to any of the datasets evaluated, and thereby an out-of-domain ruleset. This provides an example of out-of-domain performance using rules not derived from the target dataset. We observe that even when applying RBE with the Hate+Abuse ruleset we are able to outperform the baselines on each dataset. When applying RBE to new domain settings, all that is required is the authoring of additional rules for this new domain. This can be done manually, or more scalably by automatically deriving rules from the new domain data.

\begin{table*}[]
    \resizebox{\textwidth}{!}{%
    \begin{tabular}{|lllllcccccccc|}
    \hline
    \multicolumn{13}{|c|}{Content Moderation Using Rules (Unsupervised)}                                                                                                                                                                                                                                                                                                                                                                                                                                                                           \\ \hline
    \multicolumn{1}{|l|}{}                                                 & \multicolumn{4}{c|}{HateXplain}                                                                                                                       & \multicolumn{4}{c|}{Jigsaw}                                                                                                                           & \multicolumn{4}{c|}{CAD}                                                                                                                              \\ \hline
    \multicolumn{1}{|l|}{\textbf{Model}}                                   & \multicolumn{1}{c|}{Precision}      & \multicolumn{1}{c|}{Recall}         & \multicolumn{1}{c|}{F1}             & \multicolumn{1}{c|}{Acc}       & \multicolumn{1}{c|}{Precision}      & \multicolumn{1}{c|}{Recall}         & \multicolumn{1}{c|}{F1}             & \multicolumn{1}{c|}{Acc}       & \multicolumn{1}{c|}{Precision}      & \multicolumn{1}{c|}{Recall}         & \multicolumn{1}{c|}{F1}             & Acc                            \\ \hline
    \multicolumn{1}{|l|}{HateXplain Rules}                       & \multicolumn{1}{c|}{0.609}          & \multicolumn{1}{c|}{0.983}          & \multicolumn{1}{c|}{0.752}          & \multicolumn{1}{c|}{0.615}          & \multicolumn{1}{c|}{-}              & \multicolumn{1}{c|}{-}              & \multicolumn{1}{c|}{-}              & \multicolumn{1}{c|}{}               & \multicolumn{1}{c|}{-}              & \multicolumn{1}{c|}{-}              & \multicolumn{1}{c|}{-}              &                                     \\ \hline
    \multicolumn{1}{|l|}{Hate+Abuse Rules}                                 & \multicolumn{1}{c|}{0.755} & \multicolumn{1}{c|}{0.687}          & \multicolumn{1}{c|}{0.719}          & \multicolumn{1}{c|}{0.682} & \multicolumn{1}{c|}{0.164}          & \multicolumn{1}{c|}{0.361}          & \multicolumn{1}{c|}{0.226}          & \multicolumn{1}{c|}{0.972}          & \multicolumn{1}{c|}{0.586}          & \multicolumn{1}{c|}{0.193}          & \multicolumn{1}{c|}{0.290}          & \multicolumn{1}{c|}{0.909}          \\ \hline
    \multicolumn{1}{|l|}{CAD Rules}                                 & \multicolumn{1}{c|}{-} & \multicolumn{1}{c|}{-}          & \multicolumn{1}{c|}{-}          & \multicolumn{1}{c|}{-} & \multicolumn{1}{c|}{-}          & \multicolumn{1}{c|}{-}          & \multicolumn{1}{c|}{-}          & \multicolumn{1}{c|}{-}          & \multicolumn{1}{c|}{0.110}          & \multicolumn{1}{c|}{0.842}          & \multicolumn{1}{c|}{0.194}          & \multicolumn{1}{c|}{0.325}          \\ \hline
    \multicolumn{1}{|l|}{BERT$^{+\ast}$}          & \multicolumn{1}{c|}{0.606}          & \multicolumn{1}{c|}{0.990}          & \multicolumn{1}{c|}{0.752}          & \multicolumn{1}{c|}{0.613}          & \multicolumn{1}{c|}{-}              & \multicolumn{1}{c|}{-}              & \multicolumn{1}{c|}{-}              & \multicolumn{1}{c|}{-}              & \multicolumn{1}{c|}{-}              & \multicolumn{1}{c|}{-}              & \multicolumn{1}{c|}{-}              & -                                   \\ \hline
    \multicolumn{1}{|l|}{BERT$^{+\triangle}$}          & \multicolumn{1}{c|}{0.747}          & \multicolumn{1}{c|}{0.717}          & \multicolumn{1}{c|}{0.732}          & \multicolumn{1}{c|}{0.688}          & \multicolumn{1}{c|}{0.234}          & \multicolumn{1}{c|}{0.461}          & \multicolumn{1}{l|}{0.310} & \multicolumn{1}{c|}{0.977}          & \multicolumn{1}{c|}{0.587}          & \multicolumn{1}{c|}{0.205}          & \multicolumn{1}{c|}{0.303} & \multicolumn{1}{c|}{0.909}          \\ \hline
    \multicolumn{1}{|l|}{BERT$^{+\ddagger}$}          & \multicolumn{1}{c|}{-}          & \multicolumn{1}{c|}{-}          & \multicolumn{1}{c|}{-}          & \multicolumn{1}{c|}{-}          & \multicolumn{1}{c|}{-}          & \multicolumn{1}{c|}{-}          & \multicolumn{1}{l|}{-} & \multicolumn{1}{c|}{-}          & \multicolumn{1}{c|}{0.107}          & \multicolumn{1}{c|}{0.865}          & \multicolumn{1}{c|}{0.191} & \multicolumn{1}{c|}{0.290}          \\ \hline
    \multicolumn{1}{|l|}{MPNet$^{+\ast}$}          & \multicolumn{1}{c|}{0.611}          & \multicolumn{1}{c|}{0.991}          & \multicolumn{1}{c|}{0.756}          & \multicolumn{1}{c|}{0.621}          & \multicolumn{1}{c|}{-}              & \multicolumn{1}{c|}{-}              & \multicolumn{1}{c|}{-}              & \multicolumn{1}{c|}{-}              & \multicolumn{1}{c|}{-}              & \multicolumn{1}{c|}{-}              & \multicolumn{1}{c|}{-}              & -                                   \\ \hline
    \multicolumn{1}{|l|}{MPNet$^{+\triangle}$}          & \multicolumn{1}{c|}{0.652}          & \multicolumn{1}{c|}{0.850}          & \multicolumn{1}{c|}{0.738}          & \multicolumn{1}{c|}{0.641}          & \multicolumn{1}{c|}{0.247}          & \multicolumn{1}{c|}{0.501}          & \multicolumn{1}{l|}{\textbf{0.331}} & \multicolumn{1}{c|}{0.977}          & \multicolumn{1}{c|}{0.642}          & \multicolumn{1}{c|}{0.199}          & \multicolumn{1}{c|}{0.304} & \multicolumn{1}{c|}{0.912}          \\ \hline
    \multicolumn{1}{|l|}{MPNet$^{+\ddagger}$}          & \multicolumn{1}{c|}{-}          & \multicolumn{1}{c|}{-}          & \multicolumn{1}{c|}{-}          & \multicolumn{1}{c|}{-}          & \multicolumn{1}{c|}{-}          & \multicolumn{1}{c|}{-}          & \multicolumn{1}{l|}{-} & \multicolumn{1}{c|}{-}          & \multicolumn{1}{c|}{0.111}          & \multicolumn{1}{c|}{0.840}          & \multicolumn{1}{c|}{0.196} & \multicolumn{1}{c|}{0.335}          \\ \hline
    \multicolumn{1}{|l|}{Rule By Example (Distance)$^\ast$} & \multicolumn{1}{c|}{0.614}          & \multicolumn{1}{c|}{0.983}          & \multicolumn{1}{c|}{0.756}          & \multicolumn{1}{c|}{0.623}          & \multicolumn{1}{c|}{-}              & \multicolumn{1}{c|}{-}              & \multicolumn{1}{c|}{-}              & \multicolumn{1}{c|}{-}              & \multicolumn{1}{c|}{-}              & \multicolumn{1}{c|}{-}              & \multicolumn{1}{c|}{-}              & -                                   \\ \hline
    \multicolumn{1}{|l|}{Rule By Example (Distance)$^\triangle$}     & \multicolumn{1}{c|}{0.629}          & \multicolumn{1}{c|}{0.955}          & \multicolumn{1}{c|}{0.758}          & \multicolumn{1}{c|}{0.639}          & \multicolumn{1}{c|}{0.358} & \multicolumn{1}{c|}{0.284}          & \multicolumn{1}{c|}{0.317}          & \multicolumn{1}{c|}{0.986} & \multicolumn{1}{c|}{0.280} & \multicolumn{1}{c|}{0.322}          & \multicolumn{1}{c|}{0.299}          & \multicolumn{1}{c|}{0.854} \\ \hline
    \multicolumn{1}{|l|}{Rule By Example (Distance)$^\ddagger$} & \multicolumn{1}{c|}{-} & \multicolumn{1}{c|}{-}          & \multicolumn{1}{c|}{-}          & \multicolumn{1}{c|}{-}          & \multicolumn{1}{c|}{-}          & \multicolumn{1}{c|}{-}              & \multicolumn{1}{c|}{-}              & \multicolumn{1}{c|}{-}              & \multicolumn{1}{c|}{0.166}              & \multicolumn{1}{c|}{0.522}              & \multicolumn{1}{c|}{0.252}              & \multicolumn{1}{c|}{0.701}                                               \\ \hline
    \multicolumn{1}{|l|}{Rule By Example (Concat)$^\ast$}   & \multicolumn{1}{c|}{0.621}          & \multicolumn{1}{c|}{0.950}          & \multicolumn{1}{c|}{0.751}          & \multicolumn{1}{c|}{0.626}          & \multicolumn{1}{c|}{-}              & \multicolumn{1}{c|}{-}              & \multicolumn{1}{c|}{-}              & \multicolumn{1}{c|}{-}              & \multicolumn{1}{c|}{-}              & \multicolumn{1}{c|}{-}              & \multicolumn{1}{c|}{-}              & -                                   \\ \hline
    \multicolumn{1}{|l|}{Rule By Example (Concat)$^\triangle$}       & \multicolumn{1}{c|}{0.612}          & \multicolumn{1}{c|}{0.985} & \multicolumn{1}{c|}{0.755}          & \multicolumn{1}{c|}{0.621}          & \multicolumn{1}{c|}{0.189}          & \multicolumn{1}{c|}{0.052} & \multicolumn{1}{c|}{0.081}          & \multicolumn{1}{c|}{0.987}          & \multicolumn{1}{c|}{0.175}          & \multicolumn{1}{c|}{0.437}          & \multicolumn{1}{c|}{0.250}          & \multicolumn{1}{c|}{0.747}          \\ \hline
    \multicolumn{1}{|l|}{Rule By Example (Concat)$^\ddagger$} & \multicolumn{1}{c|}{-}   & \multicolumn{1}{c|}{-}          & \multicolumn{1}{c|}{-}          & \multicolumn{1}{c|}{-}          & \multicolumn{1}{c|}{-}          & \multicolumn{1}{c|}{-}              & \multicolumn{1}{c|}{-}              & \multicolumn{1}{c|}{-}              & \multicolumn{1}{c|}{0.178}              & \multicolumn{1}{c|}{0.437}              & \multicolumn{1}{c|}{0.253}              & \multicolumn{1}{c|}{0.750}                                                 \\ \hline
    \multicolumn{1}{|l|}{Rule By Example (Mean)$^\ast$}     & \multicolumn{1}{c|}{0.612}          & \multicolumn{1}{c|}{0.983}          & \multicolumn{1}{c|}{0.754}          & \multicolumn{1}{c|}{0.620}          & \multicolumn{1}{c|}{-}              & \multicolumn{1}{c|}{-}              & \multicolumn{1}{c|}{-}              & \multicolumn{1}{c|}{-}              & \multicolumn{1}{c|}{-}              & \multicolumn{1}{c|}{-}              & \multicolumn{1}{c|}{-}              & -                                   \\ \hline
    \multicolumn{1}{|l|}{Rule By Example (Mean)$^\triangle$}         & \multicolumn{1}{c|}{0.636}          & \multicolumn{1}{c|}{0.944}          & \multicolumn{1}{c|}{0.760} & \multicolumn{1}{c|}{0.646}          & \multicolumn{1}{c|}{0.188}          & \multicolumn{1}{c|}{0.124}          & \multicolumn{1}{c|}{0.149}          & \multicolumn{1}{c|}{0.984}          & \multicolumn{1}{c|}{0.294}          & \multicolumn{1}{c|}{0.273} & \multicolumn{1}{c|}{0.283}          & \multicolumn{1}{c|}{0.866}          \\ \hline
    \multicolumn{1}{|l|}{Rule By Example (Mean)$^\ddagger$}   & \multicolumn{1}{c|}{-}          & \multicolumn{1}{c|}{-}          & \multicolumn{1}{c|}{-}          & \multicolumn{1}{c|}{-}          & \multicolumn{1}{c|}{-}              & \multicolumn{1}{c|}{-}              & \multicolumn{1}{c|}{-} & \multicolumn{1}{c|}{-}             & \multicolumn{1}{c|}{0.189}              & \multicolumn{1}{c|}{0.411}              & \multicolumn{1}{c|}{0.259}              & \multicolumn{1}{c|}{0.772}                                               \\ \hline
    \multicolumn{13}{|c|}{Unsupervised Pre-Training} \\ \hline
    
    \multicolumn{1}{|l|}{Rule By Example (Mean)$^\triangle$}         & \multicolumn{1}{c|}{0.641}          & \multicolumn{1}{c|}{0.954}          & \multicolumn{1}{c|}{\textbf{0.767}} & \multicolumn{1}{c|}{0.656}                    & \multicolumn{1}{c|}{0.166}           & \multicolumn{1}{c|}{.626} & \multicolumn{1}{c|}{0.262}          & \multicolumn{1}{c|}{0.961} & \multicolumn{1}{c|}{0.260}          & \multicolumn{1}{c|}{0.320}          & \multicolumn{1}{c|}{0.287}      & \multicolumn{1}{c|}{0.846}   \\ \hline
    
    \multicolumn{1}{|l|}{Rule By Example (Distance)$^\triangle$}         & \multicolumn{1}{c|}{0.617}          & \multicolumn{1}{c|}{0.968}          & \multicolumn{1}{c|}{0.753} & \multicolumn{1}{c|}{0.624}  & \multicolumn{1}{c|}{0.203}          & \multicolumn{1}{c|}{0.465} & \multicolumn{1}{c|}{0.283}          & \multicolumn{1}{c|}{0.974}        & \multicolumn{1}{c|}{0.484}          & \multicolumn{1}{c|}{0.236}          & \multicolumn{1}{c|}{\textbf{0.317}}          & \multicolumn{1}{c|}{0.902}                    \\ \hline
    \end{tabular}%
    }
    \caption{Unsupervised Performance across all clustering strategies. $^\ast$Uses HateXplain ruleset. $^\triangle$Uses Hate+Abuse ruleset. $^\ddagger$Uses CAD ruleset. \textbf{Note:} The HateXplain Ruleset is not applicable to Jigsaw and Contextual Abuse Dataset (CAD).}
    \label{tab:unsupervised}
    \vspace{-1pc}
    \end{table*}

\subsection{Interpretability}
In addition to its improved performance, another advantage of RBE lies in its ability to perform Rule-grounding. As explained in section \ref{sec:rule_grounding}, Rule-grounding enables us to trace our model predictions back to their respective rule accompanied by the exemplars that define that rule. Table \ref{tab:rule_grounding} shows Rule-grounding examples extracted from each of our tested datasets. By nature, Rule-grounding enables two main features in RBE:

1) \textbf{Customizability/Ruleset Adaptation}: Given the vast reach of online applications, content moderation systems need to be easily adaptable to ever-emerging trends of hateful content. Particularly in online social settings, expert users of these platforms continually find new and interesting ways to bypass moderation systems. Additionally, new terminologies and slang are being introduced every day. RBE is seamlessly capable of addressing these concerns by facilitating rule-guided learning. By defining a new rule and adding at least one exemplar, RBE is able to capture emerging content without the need for re-training. Additionally, users of RBE can easily modify existing rules that may be too broad and add additional exemplars to further refine predictions in a controllable manner.

2) \textbf{Prediction Transparency}: By facilitating model interpretations via rule-grounding, users of online systems are offered tangible guidance should their content be flagged, potentially increasing user trust in the system. Additionally, this acts as a direct indicator of the type of content the rule authors want to moderate.

\subsection{Unsupervised Performance}
Table \ref{tab:unsupervised} reports our results in the unsupervised setting. We observe that RBE is able to outperform SOTA trained on noisy rules labeled samples for the HateXplain and Jigsaw dataset while also outperforming the ruleset as is on all three datasets. 
Across each dataset, we find that RBE's \textit{Distance} based strategy produces the most consistent performance, outperforming SOTA on HateXplain and CAD while performing on par with SOTA on Jigsaw. We observe that this stability in performance is due to this strategy's rule elimination objective. As opposed to the \textit{Mean} and \textit{Concat} strategies which focus on deriving rule representations in a self-supervised manner, the \textit{Distance} strategy instead focuses on eliminating over-generalized rules whose cover set of examples are semantically dissimilar. This is particularly useful in cases where precision scores are low due to a large number of false positives.

For Jigsaw, we observe a slight decrease in performance compared to SOTA. Upon further analysis, we posit that this is a result of RBE's over-reliance on the ruleset in this setting, particularly for the \textit{Mean} and \textit{Concat} strategies. This is because the ruleset directly influences the derived rule embedding due to its labeling of the cover set $C$. As such when the ruleset is over-generalized, as is the case of Hate+Abuse rules on Jigsaw, RBE is likely to match the distribution of the ruleset. We find that performing self-supervised model pre-training \cite{simcse} on the target corpus circumvents this trend for the \textit{Mean} and \textit{Concat} strategy. As such, with a more refined ruleset, a performance increase is expected as seen in HateXplain and CAD.



\section{Related Work}
There has been active work on detecting hate speech in language \cite{poletto2021resources, al2020automatic, schmidt-wiegand-2017-survey}. Hate Speech detection has proven to be a nuanced and difficult task, leading to the development of approaches and datasets targeted at various aspects of the problem \cite{cad_2021, hatexplain, MODY2023108832}. However, few attempts have been made to focus on the explainability of these models, which is an increasing area of concern surrounding their use online \cite{tarasov_2021,haimson_2021}, thus leading to the continued utilization of less powerful but more explainable methods such as rules.
Prior works have explored incorporating logical rules into model learning. \citet{awasthi_2020} proposed to weakly learn from rules by pairing them with exemplars and training a denoising model. However, this requires defining rules for all output classes, making it inapplicable to the task of hate speech detection. Additionally, this method only focuses on decreasing rule scope to solve the over-generalization problem. It does not simultaneously tackle the over-specificity problem demonstrated in Figure \ref{fig:problem}. Finally, this method does not provide a way for interpreting model predictions during inference. \citet{sunyong_2021} proposes a way to control neural network training and inference via rules, however, their framework represents rules as differentiable functions requiring complex perturbations to incorporate, making it more suitable to numerical rules such as those defined in healthcare and finance as opposed to the complex nuances of language.  \citet{reid_2022} proposes a framework for the automatic induction of symbolic rules from a small set of labeled data. However, these rules are derived from low-capacity ML models and are as a result not human-readable or explainable.

\section{Conclusion}
We introduce Rule By Example, an exemplar-based contrastive learning framework that enables learning from logical rules for accurate and explainable hate speech detection. Specifically, we propose a novel dual-encoder model architecture designed to produce meaningful rule and text representations. RBE leverages a novel exemplar-based contrastive learning objective that converges the representations of rules and text inputs of similar classes. We share results on three public datasets for hate speech detection that validate the Rule By Example framework can not only vastly outperform the initial ruleset but also outperform baseline SOTA classification methods in both supervised and unsupervised settings. Moreover, RBE enables rule-grounding which allows for more explainable model prediction benefits not available in SOTA classification methods alongside additional flexibility via Ruleset Adaptation.

\section{Limitations}
In this section, we discuss some of the limitations of the Rule by Example method.

\subsection{Dependence on Supervision}
The requirement of both a set of rules and an example per rule in our Rule by Example method means that some amount of expert supervision is required, even for the  'unsupervised' experimental setups. This could be a prohibitive cost in some scenarios. There are potential methods to select an example per rule in an unsupervised manner, such as clustering the examples the rules fires on, that could be explored in future work. However, the creation of the rules themselves means some form of expert supervision that distills knowledge about the classification task into a parseable function.

\subsection{Increased Cost Compared to Rules}
Although the Rule by Example method produces a Dual Encoder model that is shown to be much more performant than the ruleset it is derived from, it still has the cost limitations of other deep learning methods. The Dual Encoder requires far more expensive compute (GPUs) to initially train and later inference in a production setting. And even with using expensive GPUs, the latency cost is unavoidably much higher than most simple logical rules. For some applications, the quality gain of the Dual Encoder model may not be worth the increased operational cost.

\subsection{Reliance on Quality Rules and Exemplars}
Since the Rule by Example method is based on having a ruleset and associated exemplars to learn from, the quality of those rules and exemplars could affect downstream Dual Encoder model quality. If the authored ruleset and chosen exemplars are not high quality, intuitively the quality of the Dual Encoder model would suffer. This is especially true in the unsupervised setting, where the rules are used as noisy labeling functions. A possible future extension is studying the effect of rule and exemplar quality on the performance of the derived Dual Encoder model.

\section{Ethics}
Hate speech detection is a complex task. Reducing the task to authoring a set of simple logical rules can potentially lead to rule authors encoding hard biases in those rules. This can cause problems of erasure, for example, if an in-group word or an identity term is used as a rule to identify content as hate speech.

The Rule by Example method can potentially reduce these cases, for example by learning a better rule representation and identifying when a term is used as in-group speech as opposed to being used as an insult or slur. However, the derived Dual Encoder is also at the risk of propagating and amplifying these biases \cite{hall}, causing greater unintended harm than the original ruleset. 

Whether using a ruleset or using a more complicated model, it is important to support classifiers with additional Responsible AI work streams, such as reviews of classifier behavior and measurements of fairness.

\section*{Acknowledgements}
We thank our anonymous reviewers for their feedback and suggestions. This work was conducted by the ROAR (Responsible \& Open AI Research) team at Microsoft Cloud \& AI. At UofM, Christopher Clarke is supported in part by award NSF1539011 by the National Science Foundation.

\bibliography{anthology,custom}
\bibliographystyle{acl_natbib}

\appendix



\end{document}